\documentclass[pdflatex,sn-nature,Numbered]{sn-jnl}

\usepackage{graphicx}
\usepackage{multirow}
\usepackage{amsmath,amssymb,amsfonts}
\usepackage{booktabs}
\usepackage{xcolor}

\newcommand{\bag}{B}                          
\newcommand{\cand}{Y}                         
\newcommand{\proto}[1]{\boldsymbol{\mu}_{#1}} 
\newcommand{\pseudo}{\mathbf{s}}              
\newcommand{\nlev}{K}                         
\newcommand{\naxes}{E}                        

\begin{document}
\title{Surg-Pico: Disentangling Granular Surgical Skills from Video-Level Labels}

\author*[1]{\fnm{Julien} \sur{Quarez}}\email{julien.quarez@kcl.ac.uk}
\author[1]{\fnm{Marc} \sur{Modat}}
\author[1]{\fnm{Sebastien} \sur{Ourselin}}
\author[1,2]{\fnm{Jonathan} \sur{Shapey}}
\author[1]{\fnm{Alejandro} \sur{Granados}}

\affil[1]{\orgdiv{School of Biomedical Engineering and Imaging Sciences},
\orgname{King's College London}, \orgaddress{\city{London}, \country{UK}}}
\affil[2]{\orgdiv{Neurosurgery}, \orgname{King's College Hospital},
\orgaddress{\city{London}, \country{UK}}}

\abstract{
Performance is mostly assessed at a few key moments across an exercise; despite this, deep-learning approaches to surgical skill assessment typically regress a single score from an entire video.
This is further complicated by low inter-rater agreement, making a global score a noisy and ambiguous supervision target. We introduce Surg-Pico, a multiple-instance partial-label learning framework that addresses both challenges. Building on PiCO+, Surg-Pico assigns skill levels to individual clips, separating those consistent with video-level labels from those affected by label ambiguity. On JIGSAWS, Surg-Pico achieves a mean Spearman correlation of 0.69 across OSATS criteria, outperforming the best previous method by 0.07. On AIxSuture, it reaches 83\% of the performance ceiling imposed by inter-rater reliability on the GRS. Beyond its improved predictive performance, Surg-Pico provides interpretable temporal skill profiles, suggesting that skill-related signals are concentrated in critical surgical gestures.}

\maketitle

\begin{figure}[t]
    {\includegraphics[width=0.965\linewidth]{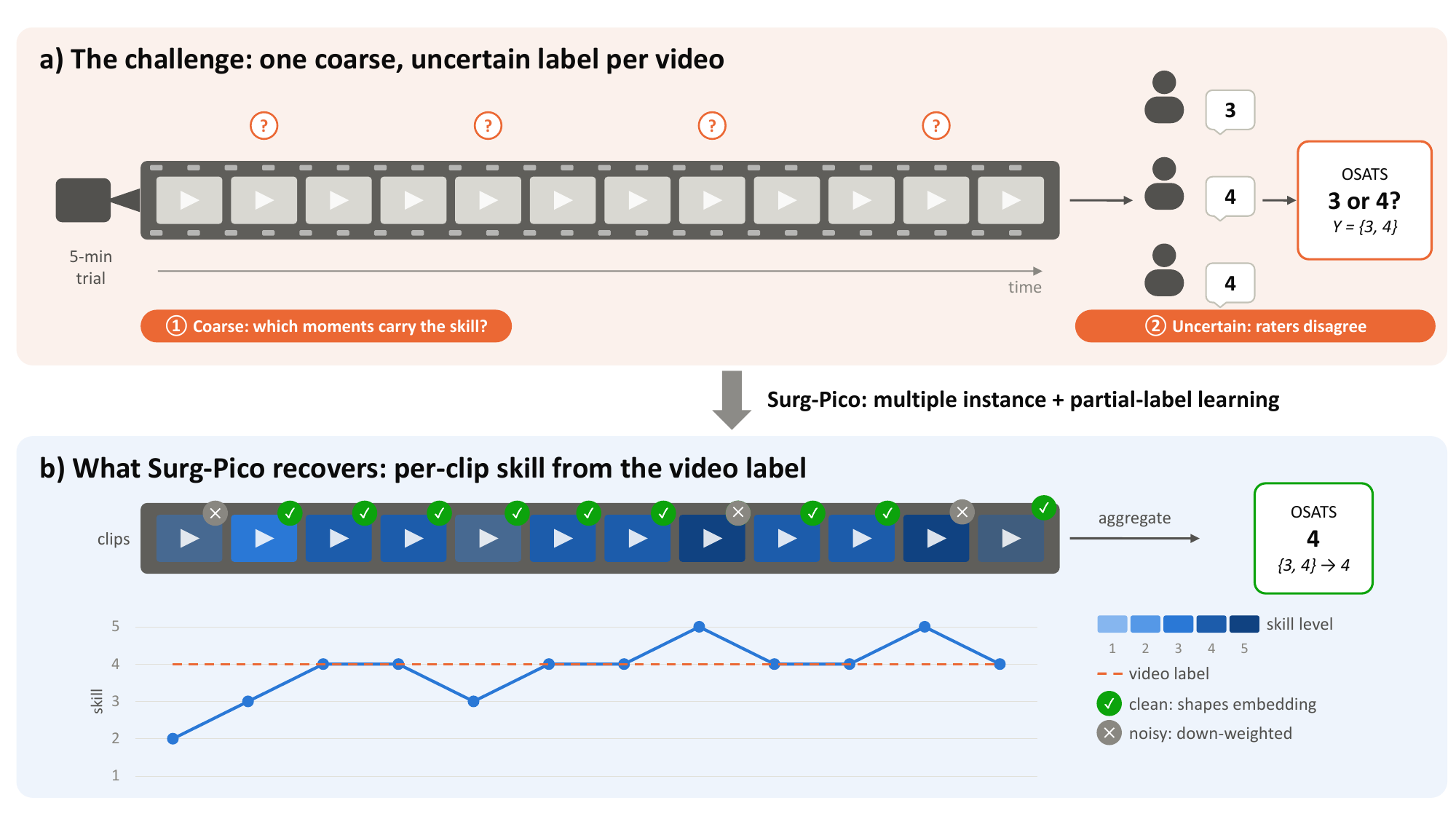}}
    \caption{The challenge and our approach. \textbf{a}, A trial is a long
    video that receives a global OSATS score, so the moments that
    carry the skill are unknown (coarse). Furthermore, raters often disagree,
    leaving a set of plausible levels rather than one label (uncertain).
    \textbf{b}, Surg-Pico treats the video as a set of non-overlapping clips and the rater disagreement as a candidate set. It recovers a skill level for every clip, keeps
    clips consistent with their label (clean) to shape the representation, and
    down-weights the rest (noisy). This allows the model to resolve the video-level label. Per-clip
    values are illustrative.}
    \label{fig:concept}
\end{figure}

\section{Introduction}\label{sec:introduction}

The World Health Organization estimates a shortfall of 11 million health workers by 2030 (\url{https://www.who.int/health-topics/health-workforce}). In Italy alone, 37\% of general surgery residency positions were left vacant in 2025~\cite{Cavallaro2026-tb}. While this shortfall is multifactorial, recent evidence suggests that surgical training has not kept up with demand~\cite{Kashkosh2025-md,bma2020consultant}, and that limited exposure to certain surgical cases has compounded the issue~\cite{killoran2024open}, with consultant training lagging~\cite{srinivasa2025consultant}.
Limited exposure means that residents often rely on bench-top, phantom and cadaver-based training as surrogates for operating-room experience and as opportunities to develop technical skills~\cite{sutherland2006surgical,killoran2024open}. However, these training environments remain constrained by the availability of senior surgeons who can provide expert feedback~\cite{godbole2026automated}. Artificial intelligence (AI)-based assessment has long been proposed as a remedy~\cite{yanik2021deep, godbole2026automated}, increasing training opportunities while reducing the dependence on expert supervision.

Automated surgical skill assessment, however, is hardest at the granularity at
which it would be the most useful~\cite{Lam2022,haque2022ease, OSSChallenge,sirajudeen2024deep}. While a trainee may receive a single overall
performance score for a procedure, useful feedback needs to identify the
specific moments at which technical performance was competent or
deficient~\cite{Ende1983feedback,Bonrath2015coaching,Kelly2019,MARKUS}. Obtaining ground truth at that level of
granularity is fundamental to the translation of AI tools into surgical
training~\cite{maierhein2022surgical}. In practice, manually annotating progressively shorter temporal
segments does not scale and introduces inconsistencies: as segments
shrink, workload increases and disagreement grows, making expert raters'
judgement less reliable~\cite{Meireles2021,maierhein2022surgical}. This limitation is reflected in the available
open-source datasets. 

Most work has relied on JIGSAWS~\cite{jigsaw}, in which
each bench-top exercise is rated at the video level by a single expert. AIxSuture~\cite{Hoffmann2024} has a similar structure, but each video is rated
by three independent raters; the Open Suturing Skills (OSS)
challenge~\cite{OSSChallenge} later extended it with further recordings that
remain private. 

Inconsistencies in labelling also introduce ambiguity into the supervision
signal. In AIxSuture, the average pairwise Pearson
correlation between raters is about $0.8$, and single-rater reliability ranges
from $0.54$ to $0.81$ across assessment categories (Supplementary Note~1). This
suggests that single-rater labels such as those of JIGSAWS carry substantial
noise. Supervision for surgical skill assessment is therefore not only
coarse but also ambiguous.

Earlier work has regressed the global rating score (GRS) or expertise level from
bench-top videos~\cite{wang2020towards,li2022visa,Hoffmann2024,guo2024mtvit}.
Methods have since shifted towards predicting the individual Objective
Structured Assessment of Technical Skills (OSATS)~\cite{Martin1997} sub-scores,
to make assessment more interpretable~\cite{benmansour2023deep,recap,OSSChallenge}.
Several works also model performance over time. Some learn segment
representations by comparing trials: against a gold-standard
reference~\cite{anastasiou2023keep}, or through attention streams that
accumulate evidence of high and low skill under a ranking
objective~\cite{doughty2019pros}. Others weight or aggregate segments within a
trial~\cite{wang2020towards,liu2021towards,li2022visa}, or localise
skill-related events within long videos~\cite{ding2023sedskill}. Closest to
this study, ReCAP~\cite{recap} recursively tracked performance across a session
and produced segment-level OSATS pseudo-labels at no extra annotation cost. In
all of these, temporal structure is inferred rather than directly supervised:
segment representations are shaped only by a video-level objective. Where they
are learned by comparing trials~\cite{anastasiou2023keep,who}, they amplify the differences between surgeons.
Uninformative or atypical parts of a trial can therefore exert
disproportionate influence on the learned representation, and there is no
signal at training time to correct it.

Two established lines of work partly address these problems. Multiple
instance learning (MIL) is a form of weak supervision in which a label is
available only for a bag of instances, and the model must recover the appropriate instance latent structure from that bag label alone. First developed for
drug-activity prediction~\cite{Dietterich1997}, it is now widely used for
whole-slide images, where the evidence for a slide-level diagnosis lies in a
small subset of patches~\cite{Ilse2018,Qu2023RethinkingMIL,Zhang2022-mq,Sun2025-ni}.
Instance-based methods propagate the bag label to every instance, which injects
label noise when few instances carry the evidence~\cite{Carbonneau2018}.
Bag-based methods pool instances into one representation and achieve higher
bag-level accuracy~\cite{Wang2018}, at the cost of instance-level
interpretability~\cite{Jain2019,Javed2022}. Weakly supervised video anomaly
detection~\cite{Sultani2018} and temporal action
localisation~\cite{10671581} also recover segment-level evidence from a video
label, but their supervision is categorical and one-sided. Skill, in contrast,
is continuous across time, and the label of each segment is ambiguous rather
than merely present or absent. Partial-label learning (PLL) formalises this
setting: each sample carries a set of candidate labels of which one is
correct, and the model progressively disambiguates the ground truth during
training~\cite{Cour2011,Feng2020,Lv2020-zr,wang2025realistic}.
PiCO~\cite{Wang2022PiCO} does so by combining contrastive representation
learning with class prototypes. PiCO+~\cite{Wang2022PiCOPlus} goes one step
further and addresses examples whose true label lies outside the candidate
set, supervising clean and corrupted samples differently.

The two ambiguities are distinct: one asks which instance carries the
evidence, the other which level that evidence corresponds to. Their
combination has been formalised as multiple-instance partial-label learning
(MIPL)~\cite{Tang2024MIPL,Tang2023DEMIPL}. We argue that surgical skill
assessment presents both problems at once (Fig.~\ref{fig:concept}) and is
therefore well suited to MIPL.

Closest to our approach, Qu et al.~\cite{Qu2023RethinkingMIL} adapted PiCO to whole-slide image MIL. To our knowledge, PiCO+, has not been applied to MIL. Surg-Pico does so for ordinal skill levels, with candidate sets derived from rater disagreement, a per-bag clean/noisy split and a bag-level centroid read-out.
Our contributions are:

\begin{itemize}
\item \textbf{Surgical skill assessment formulated as multiple-instance partial-label learning.} We formulate a video as a bag of distinct temporal segments with skill levels that are unobserved and are only constrained by the OSATS video label.
  The model is designed to resolve the mismatch between the assigned video label and the underlying clip scores.

\item \textbf{Surg-Pico, a bag-level adaptation of PiCO+.}
We extend PiCO+'s core components to video inputs, where a video centroid is decoded against a skill prototype bank, allowing for both video supervision and clip disambiguation.

\item \textbf{Per-criterion evaluation on two surgical datasets.} We report state-of-the-art results on
  JIGSAWS~\cite{jigsaw} and the first per-criterion results on the public AIxSuture~\cite{Hoffmann2024} data, while including performance as a fraction
  of the attenuation bound set by inter-rater reliability on AIxSuture.
\end{itemize}

\section{Results}\label{sec:results}
\subsection{Surg-Pico overview}\label{sec:results-overview}
\begin{figure}[htbp]
{\includegraphics[width=\textwidth]{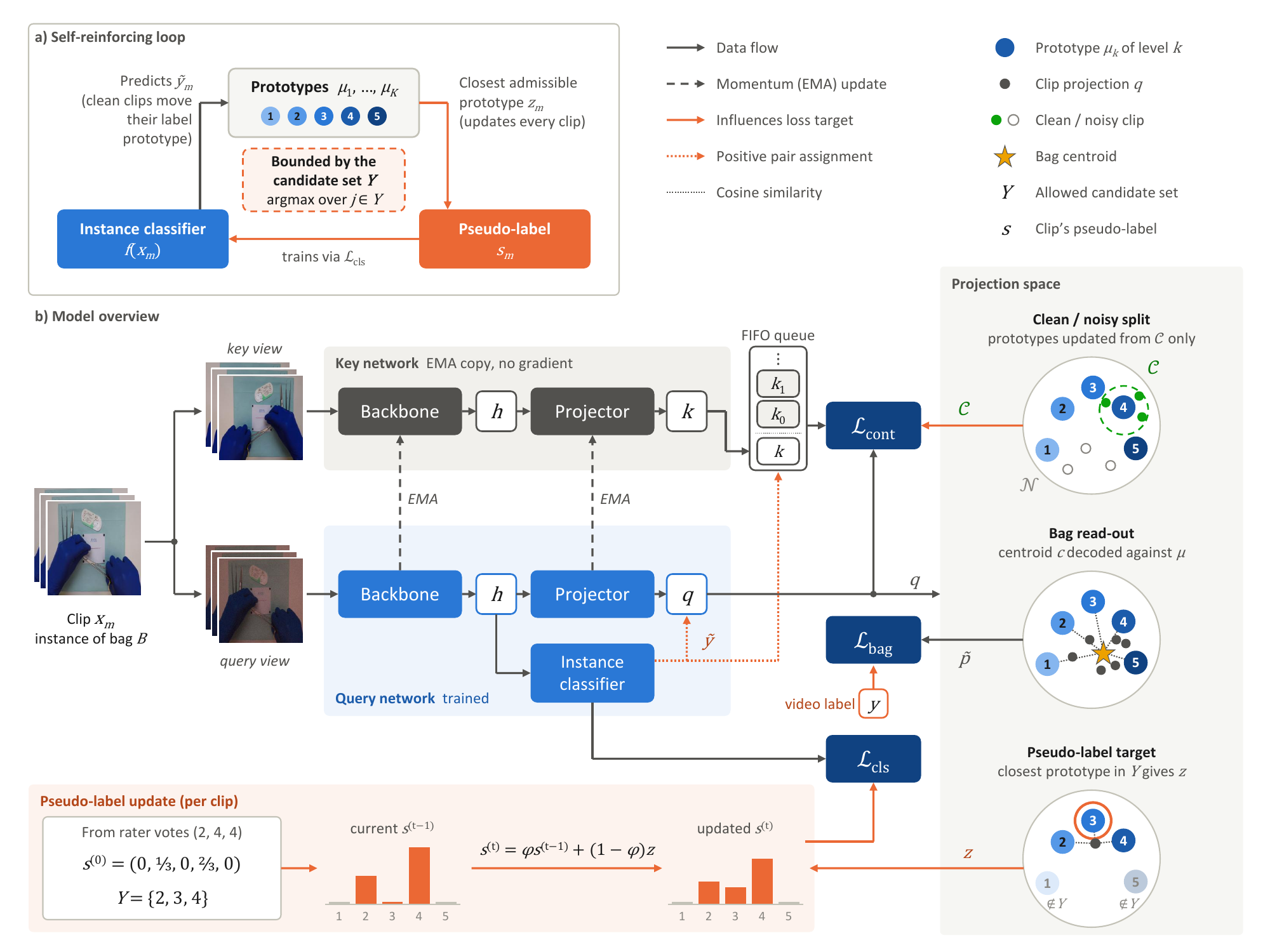}}
    \caption{Overview of Surg-Pico. \textbf{a}, Self-reinforcing loop between the
    instance classifier, the prototypes and the pseudo-labels. \textbf{b}, Model
    overview. The pseudo-label of each clip (instance),
    $s^{(0)}$, is initialised from the rater vote proportions over the $K$
    levels and may move anywhere within the hull of those votes, $Y$. Each clip
    is augmented differently for the query and key paths. The instance classifier is trained against the running
    pseudo-label, and the prototype space is used to update both the
    pseudo-labels and the prototypes. A final loss is applied to the centroid of
    the projected clips of a bag against the video label. The losses are defined in Sec.~\ref{sec:bag-loss} and Supplementary Note~2.}
    \label{fig:architecture}
\end{figure}

Surg-Pico treats each video as a bag of non-overlapping 5-s clips, and each
OSATS criterion as a separate partial-label problem (Fig.~\ref{fig:architecture};
Methods).

Each clip receives a pseudo-label initialised from the raters' votes
and restricted to a candidate set. On JIGSAWS this set is the video label; on
AIxSuture it is the range spanned by the three raters' scores. A bank of
prototypes, one per skill level, is updated by a proportion $\delta$ of instances that the model considers clean ($\delta=0.5$ on JIGSAWS and 0.8 on AIxSuture). The remaining noisy clips are supervised differently, allowing them to depart from their video label. A video's score is read
from the centroid of its clips' projections against the prototypes.

We evaluate on JIGSAWS (103 robot-assisted trials, one rater, six criteria)
under leave-one-supertrial-out (LOSO) and leave-one-user-out (LOUO)
cross-validation. We also evaluate on AIxSuture (314 open-suturing videos,
three raters, eight criteria), using the official split and a 4-fold participant-disjoint cross-validation (leave-multiple-users-out, the counterpart of LOUO), so that no participant appears in both training and test folds. We report Spearman's $\rho$ per criterion and averaged over criteria
($\rho_{\mathrm{OSATS}}$), plus macro-F1 and expected cost (EC) for comparison
with the OSS challenge. GRS is built from the OSATS predictions, and not regressed. On AIxSuture, each correlation is compared with the
ceiling set by inter-rater reliability, $\sqrt{\mathrm{ICC}}$. All results are
means over three seeds.

\subsection{Surg-Pico improves on prior JIGSAWS results}\label{sec:results-jigsaws}
Our method outperforms the best available previous results on LOSO, Table~\ref{tab:osats-average}, by 7 points against ReCAP trained jointly on the three tasks
(0.62), and by 8 points against FCN on their task mean.
Table~\ref{tab:osats-specific} reports per-criterion Spearman's $\rho$ under both
protocols. Under LOSO, Surg-Pico improves on ReCAP on every criterion, from 0.01 on suture and needle handling to 0.13 on time and motion. Under LOUO, the mean falls to 0.38, ranging from
0.18 (suture and needle handling) to 0.48 (time and motion).
\begin{table}[h]
\centering
\caption{Per-criterion Spearman's $\rho$ on JIGSAWS, across tasks (mean $\pm$ SD
over seeds). RT: respect for tissue; TM: time and motion; OP: overall
performance; SNH: suture and needle handling; FO: flow of operation; QFP:
quality of final product.}
\label{tab:osats-specific}
\begin{tabular}{lccc}
\toprule
 & ReCAP~\cite{recap} & \multicolumn{2}{c}{Surg-Pico} \\
\cmidrule(lr){2-2}\cmidrule(lr){3-4}
Criterion & LOSO & LOSO & LOUO \\
\midrule
RT   & 0.56 & \textbf{0.65} $\pm$ 0.02 & 0.45 $\pm$ 0.01\\
TM   & 0.62 & \textbf{0.75} $\pm$ 0.01 & 0.48 $\pm$ 0.01\\
OP   & 0.65 & \textbf{0.68} $\pm$ 0.02 & 0.38 $\pm$ 0.01\\
SNH  & 0.65 & \textbf{0.66} $\pm$ 0.02 & 0.18 $\pm$ 0.10\\
FO   & 0.64 & \textbf{0.74} $\pm$ 0.01 & 0.43 $\pm$ 0.04\\
QFP  & 0.62 & \textbf{0.66} $\pm$ 0.01 & 0.31 $\pm$ 0.05\\
\midrule
Mean & 0.62 & \textbf{0.69} $\pm$ 0.01 & 0.37 $\pm$ 0.03 \\
\bottomrule
\end{tabular}
\end{table}

Most of this
gap reflects what the two protocols measure. Under LOSO, every surgeon
contributes trials to both training and test folds, so identity is itself
predictive. Prototypes are vulnerable to this. With eight
surgeons, each level's prototype is estimated from the clips of only a few
people, so it reflects their individual style as well as their skill, which does not transfer well to an unseen surgeon under LOUO.

\begin{table}[t]
\centering
\caption{$\rho_{\mathrm{OSATS}}$ on JIGSAWS, averaged over the six criteria
under LOSO, per task (KT: knot tying; NP: needle passing; SU: suturing) and
across tasks (AT). *: trained jointly on the three tasks. K: kinematic data. V: video data.}
\label{tab:osats-average}
\begin{tabular}{lcccc}
\toprule
Method & KT & NP & SU & AT \\
\midrule
APEN (K)~\cite{Zia2018}         & 0.66          & 0.45          & 0.59          & 0.57 \\
FCN (K)~\cite{IsmailFawaz2019}  & 0.65          & \textbf{0.57} & 0.60          & 0.61 \\
ReCAP (K)~\cite{recap}          & 0.70 & 0.46          & 0.62 & 0.59 / 0.62* \\
Surg-Pico (V) (ours)            & \textbf{0.75}*            & 0.53*            & \textbf{0.68}*            & \textbf{0.69}* \\
\bottomrule
\end{tabular}
\end{table}

\subsection{Surg-Pico reaches 77-86\% of the inter-rater reliability ceiling on AIxSuture}\label{sec:results-aix}

\begin{table}[h]
\centering
\caption{Results on AIxSuture on the official test split (labels capped at 4 due to absence of label 5 in train folds for certain criteria)
and under 4-fold cross-validation (CV, 3-seed mean). Cross-validated
performance is also given as a fraction of the attenuation bound set by each
criterion's inter-rater reliability: $\mathrm{ICC}(C,3)$ is the three-rater
consistency intraclass correlation (Supplementary Table~1), and the bound on an
observed correlation against a target of that reliability is
$\sqrt{\mathrm{ICC}}$.}
\label{tab:aix-main}
\begin{tabular}{lcccc}
\toprule
Metric & Test (capped at 4) & CV & $\sqrt{\mathrm{ICC}}$ & \% of bound \\
\midrule
$\rho_{\mathrm{OSATS}}$ (eight criteria) & 0.71 $\pm$ $<$0.01 & 0.77 $\pm$ $<$0.01 & -- & -- \\
GRS $\rho$                         & 0.74 $\pm$ $<$0.01 & 0.80 $\pm$ $<$0.01 & 0.96 & 83 \\
Macro-F1                           & 0.44 $\pm$ $<$0.01 & 0.43 $\pm$ $<$0.01 & -- & -- \\
EC $\downarrow$                                 & 0.18 $\pm$ 0.01 & 0.13 $\pm$ 0.01 & -- & -- \\
\midrule
\multicolumn{5}{l}{\emph{Per-criterion $\rho$}} \\
Respect for Tissue      & 0.71 $\pm$ $<$0.01 & 0.68 $\pm$ 0.01 & 0.88 & 77 \\
Time and Motion         & 0.65 $\pm$ $<$0.01 & 0.77 $\pm$ 0.01 & 0.92 & 84 \\
Instrument Handling     & 0.72 $\pm$ $<$0.01 & 0.79 $\pm$ 0.01 & 0.93 & 85 \\
Suture Technique        & 0.71 $\pm$ 0.01 & 0.77 $\pm$ $<$0.01 & 0.93 & 83 \\
Flow of Operation       & 0.73 $\pm$ 0.02 & 0.79 $\pm$ 0.01 & 0.92 & 86 \\
Knowledge of Procedure  & 0.74 $\pm$ 0.01 & 0.80 $\pm$ $<$0.01 & 0.96 & 83 \\
Overall Performance     & 0.70 $\pm$ $<$0.01 & 0.80 $\pm$ $<$0.01 & 0.95 & 84 \\
Final Quality           & 0.72 $\pm$ 0.01 & 0.77 $\pm$ $<$0.01 & 0.95 & 81 \\
\bottomrule
\end{tabular}
\end{table}

Table~\ref{tab:aix-main} shows that the model approaches the reliability ceiling for a number of OSATS categories. There is a gap of 0.12 between the
best (knowledge of procedure) and worst (respect for tissue) criteria, whereas
the remaining criteria vary by only about 0.01 (SD).
The $\sqrt{\mathrm{ICC}}$ column puts this gap in perspective: respect for tissue is
the criterion on which raters disagree the most, which lowers the maximum
achievable $\rho$ (Eq.~\eqref{eq:attenuation}). We only loosely compare these
results with the published metrics of the OSS challenge
(Table~\ref{tab:challenge-context}), as the challenge reports performance on a
test set to which we do not have access. Our model falls within the range reported by the challenge.

\begin{table}[t]
\centering
\caption{Comparison with the results of the OSS
challenge~\cite{OSSChallenge}. BSE: best submitted entry.}
\label{tab:challenge-context}
\begin{tabular}{llcc}
\toprule
Evaluation data & Method & Macro-F1 $\uparrow$ & EC $\downarrow$ \\
\midrule
\multirow{2}{*}{OSS base}
  & Organisers' baseline & 0.39 & 0.16 \\
  & BSE                  & 0.38 & 0.16 \\
\midrule
\multirow{3}{*}{OSS + expert videos}
  & Organisers' baseline & 0.47 & 0.16 \\
  & BSE 1                & 0.30 & 0.21 \\
  & BSE 2                & 0.29 & 0.20 \\
\midrule
AIxSuture, 4-fold CV & Surg-Pico (ours) & 0.43 & 0.13 \\
\bottomrule
\end{tabular}
\end{table}

\subsection{Contrastive prototypes and releasing clips from the video label drive the gains}\label{sec:results-ablations}
To assess how the different components of the method affect performance, we vary the ``read-out'', ``PLL module'' and ``candidate set'' and report the results in Table~\ref{tab:ablation}. With regard to read-out, we adapt the ReCAP prediction head to our backbone; see Supplementary Note~3. The candidate set varies on AIxSuture by either the range of available rater labels or, similarly to JIGSAWS, being limited to the single video label. The PLL module is either absent (each clip is trained on its video label, i.e. instance-level MIL), PiCO (no clean/noisy split) or PiCO+ (clean/noisy split).
\begin{table}[t]
\centering
\caption{Ablation over candidate set, read-out and PLL module.
$\rho_{\mathrm{OSATS}}$ and macro-F1 under LOSO on JIGSAWS and under 4-fold CV
on AIxSuture (mean $\pm$ SD). The rater hull
applies to AIxSuture only, as JIGSAWS has a single rater
(Sec.~\ref{sec:candidate}).}
\label{tab:ablation}
\begin{tabular}{llcccc}
\toprule
 & & \multicolumn{2}{c}{JIGSAWS} & \multicolumn{2}{c}{AIxSuture} \\
\cmidrule(lr){3-4}\cmidrule(lr){5-6}
Read-out & PLL module & $\rho_{\mathrm{OSATS}}$ & Macro-F1 & $\rho_{\mathrm{OSATS}}$ & Macro-F1 \\
\midrule
\multicolumn{6}{l}{\textit{collapsed candidate set}} \\
ReCAP    & None  & 0.62 $\pm$ 0.01   &  0.36 $\pm$ 0.02     & 0.70 $\pm$ 0.01 & 0.34 $\pm$ 0.01  \\
ReCAP    & PiCO  & 0.66 $\pm$ 0.01 & 0.37 $\pm$ 0.01         & 0.74 $\pm$ 0.01          & 0.40 $\pm$ $<$0.01 \\
ReCAP    & PiCO+ & 0.68 $\pm$ 0.01 &  0.38 $\pm$ $<$0.01 & 0.76 $\pm$ 0.01          & 0.42 $\pm$ $<$0.01 \\
Centroid & PiCO+ & \textbf{0.69} $\pm$ 0.01 & \textbf{0.41} $\pm$ 0.02 & 0.76 $\pm$ 0.01          & 0.42 $\pm$ 0.01 \\
\midrule
\multicolumn{6}{l}
{\textit{rater-hull candidate set}} \\
Centroid & PiCO  & n/a &          & 0.76 $\pm$ $<$0.01          & 0.42 $\pm$ $<$0.01 \\
Centroid & PiCO+ & n/a  &         & \textbf{0.77} $\pm$ $<$0.01 & \textbf{0.43} $\pm$ $<$0.01 \\
\bottomrule
\end{tabular}
\end{table}

Replacing the plain classifier (instance-level MIL) with PiCO gives
the largest single gain, raising $\rho_{\mathrm{OSATS}}$ from 0.62 to 0.66 on
JIGSAWS and from 0.70 to 0.74 on AIxSuture. The baseline without a PLL module matches published ReCAP on JIGSAWS (0.62),
so the 0.07 that Surg-Pico adds over it mirrors its gain over ReCAP in
Table~\ref{tab:osats-average}.

With a collapsed candidate set, PiCO
pins every clip to the label of its video. This configuration is therefore the
reference against which to judge any mechanism that lets a clip depart from that
label. Surg-Pico offers two such mechanisms: the noisy set of PiCO+, whose
target spans all $\nlev$ levels, and the rater hull, which widens the candidate
set itself. On JIGSAWS, where only the first is available, it raises
$\rho_{\mathrm{OSATS}}$ from 0.66 to 0.68. On AIxSuture, either mechanism alone
raises it from 0.74 to 0.76 and macro-F1 from 0.40 to 0.42, and the two
together reach 0.77 and 0.43. These results suggest that what matters is that clips are not bound
to the video label, more than how they are released from it.

The centroid read-out adds 0.01 in
$\rho_{\mathrm{OSATS}}$ and 0.03 in macro-F1 on JIGSAWS and leaves AIxSuture unchanged
under PiCO+ (0.76 and 0.42 with either read-out), which also indicates that the
AIxSuture gains above are not a read-out effect. A possible explanation lies in
where clips are aggregated. The ReCAP read-out averages per-clip class
probabilities, so an overconfident but uninformative clip pulls the video
prediction as much as an informative one. The centroid read-out averages
projections, so the influence of a clip is bounded by its position in the
projection space. We would expect this to matter most on a small dataset such
as JIGSAWS, where a few such clips carry more weight.

\subsection{Clean-clip selection favours the core gestures of each task}\label{sec:results-gestures}
\begin{figure}[t]
{\includegraphics[width=\textwidth]{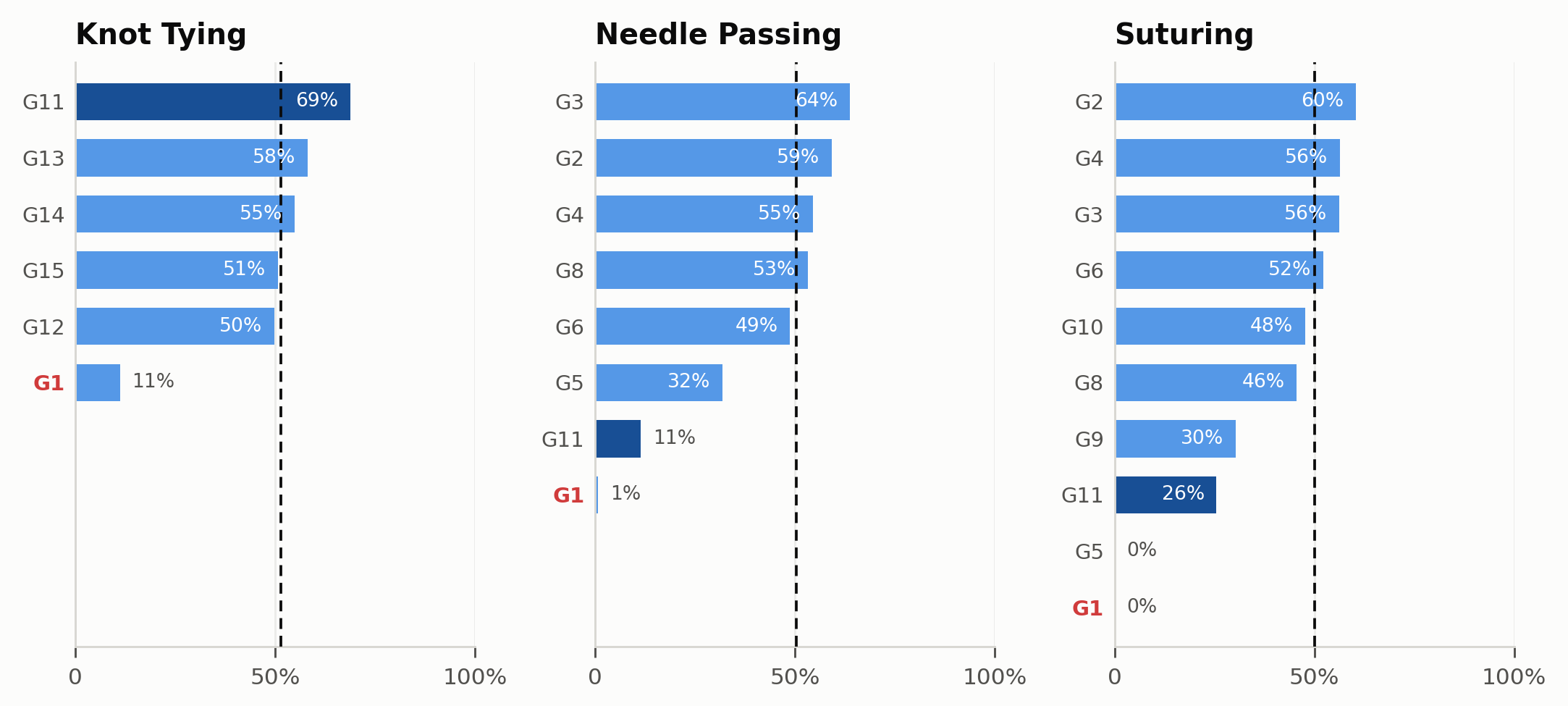}}
\caption{Fraction of clips retained in the clean set $\mathcal{C}$ per gesture,
for the three JIGSAWS tasks, on the overall performance criterion. The dashed line marks
$\delta = 50\%$.
G1: reach for needle (right hand); G2: position needle; G3: push needle through
tissue; G4: transfer needle from left to right; G5: move to centre with needle
in grip; G6: pull suture (left hand); G8: orient needle; G9: tighten suture
with right hand; G10: loosen more suture; G11: drop suture and move to end
points; G12: reach for needle (left hand); G13: make C-loop around right hand;
G14: reach for suture (right hand); G15: pull suture with both hands.}
    \label{fig:gestures}
\end{figure}

JIGSAWS provides gesture annotations for each exercise. We record which
clips were retained as clean, i.e.\ assigned to $\mathcal{C}$, during training,
and visualise them as a ratio over the gesture annotations in
Fig.~\ref{fig:gestures}, where by construction, mean retention equals $\delta= 50\%$. Since $\delta$ fixes how many clips are retained, not which gestures they belong to, retention above 50\% indicates a preference for that gesture. A clip is assigned to a gesture if at least 80\% of its frames fall inside the annotated range. 

Qualitatively, the selection appears to target the
gestures that are most informative for skill assessment. In needle passing and
suturing, the model favours G2, G3 and G4 (55-64\%),
which correspond to critical elements of these exercises. In contrast, G1 is
rarely, if ever, retained (0-11\%): as the starting gesture of every exercise,
it carries the least information about overall performance. We also observe a
difference in the retention of G11 across tasks, which suggests that the
selection is specific to the task, if not to individual videos. Needle passing
and suturing both end in states (G11) that look very different from the rest
of the exercise, whereas the end of knot tying closely resembles the movements
performed throughout.

\subsection{Clip-level scores order skill levels but partly track suture throughput}\label{sec:results-clip}
\begin{figure}[t]
    \centering
    \includegraphics[width=\textwidth]{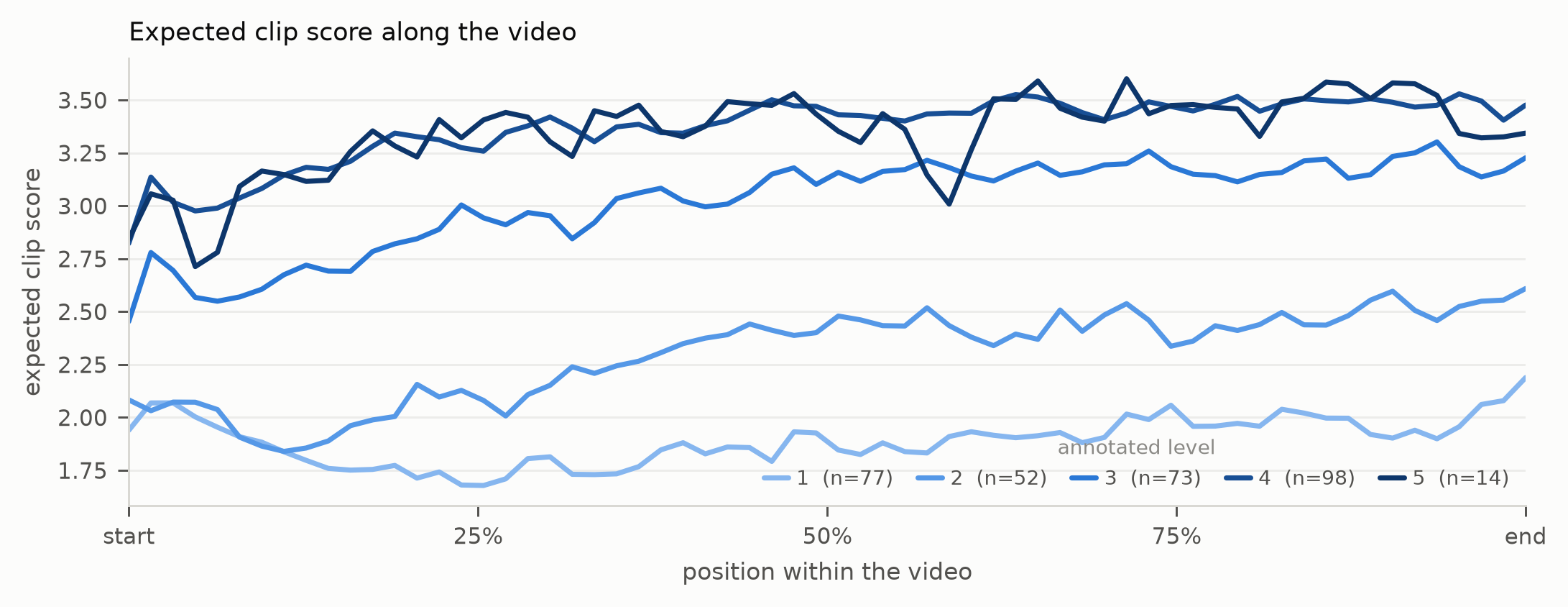}
    \caption{Expected clip score across held-out folds for the knowledge of
    procedure criterion (highest ICC) on AIxSuture. Each trial is resampled onto a common
    normalised position axis and averaged within its annotated level.}
    \label{fig:clip}
\end{figure}
We record the expected level of each clip for each held-out fold and visualise
them in Fig.~\ref{fig:clip}.    Videos labelled 4 and 5 are hardly
distinguishable, and the expected level increases over the course of each
video.
Both observations follow from the data. Levels 4 and 5 are hard to separate because only 14 videos (4.5\%) carry a
level-5 label, and in 93\% of them at least one rater scored 4. Very few
samples can therefore anchor the prototype $\proto{5}$, and the method can pull
most of them towards $\proto{4}$.

We hypothesise that the positional increase stems from how the videos were
recorded: within a time limit, participants were asked to perform as many
sutures as possible. Suture count, available as metadata, correlates with the
GRS of each video at $\rho = 0.85$. This suggests that the model partly
lets the number of visible sutures affect its predictions, i.e.\ more visible
sutures, higher predicted performance.
The effect is nonetheless limited:
clip position within a video explains only 2.3\% of the clip-score variance, and predictions still correlate with the rated score after
controlling for suture count (partial $\rho = 0.47$, $\rho$(prediction, GRS) = 0.80, $\rho$(prediction, sutures) = 0.74, $\rho$(sutures, GRS) = 0.85; $n = 314$).

\section{Discussion}\label{sec:discussion}
Previous formulations, such as ReCAP~\cite{recap}, treat every clip as an equally valid observation of a trial's skill level. At a fine enough temporal resolution, this assumption becomes harmful: the proportion of informative clips decreases, reducing the signal-to-noise ratio and allowing uninformative samples to introduce unwanted bias into the learned representation. This is shown on both datasets, where propagating the bag-level label to every clip allows noisy or uninformative instances to influence the embedding space and adversely affect performance.

This observation is consistent with the nature of surgical skill assessment, where assessable elements are often sparse within a video. Our approach allows an informed guess on how much a video presents a pertinent signal to optimise against.
This more closely reflects clinical assessment, in which a clinician may base an overall judgement on a small number of specific moments rather than evaluating every moment equally. This distinction is particularly relevant for longer exercises, where a substantial proportion of a trial may not directly reflect the assigned performance label. The problem is further compounded by inter-rater variability, which makes instance-level ground truth inherently ambiguous.

Inter-rater reliability, measured by the ICC, also provides an informative upper bound on performance.
This is particularly important when developing models intended to generalise across surgeons and assessment settings. Aggregating assessments from multiple experts provides a more robust target, but does not eliminate subjectivity. Evaluating model performance relative to inter-rater reliability therefore provides a more meaningful indication of how effectively a model captures the underlying assessment signal.

Beyond predictive performance, the proposed disambiguation framework offers opportunities for interpreting the temporal distribution of skill-related information. Through contrastive learning, the model can move beyond assigning relatively uniform performance estimates across clips and instead distinguish between individual components of an exercise. Although no instance-level ground truth is available to establish whether the resulting pseudo-labels are correct, these predictions provide an additional means of auditing model behaviour and guiding subsequent development.

For example, assigning a high skill level to a clip that contains little relevant information may indicate that the model has learned an inappropriate representation. Conversely, the retention of meaningful surgical gestures in JIGSAWS provides encouraging evidence that the model identifies temporally localised signals associated with skill. This offers an opportunity to investigate which gestures contribute most strongly to the final assessment. An interesting extension would be to explicitly model uninformative instances and prevent them from contributing to the final prediction altogether, rather than relying solely on their reduced influence during representation learning.

An important limitation, however, is that $\delta$ is set statically across the dataset. A more principled approach would allow $\delta$ to vary at the bag level, adapting to the amount of relevant information present in each trial. We leave this as an important direction for future work.
Another limitation is the size of the JIGSAWS corpus. With eight
surgeons, LOSO leaves surgeon identity available to any model, so LOSO
results, ours and previously published ones, are best read as a benchmark
comparison rather than as evidence of generalisation. LOUO removes this
shortcut but rests on eight held-out surgeons, so its estimate is imprecise,
and prototypes learned from so few people cannot fully separate skill from
individual style. Nonetheless, by framing the problem as instance selection on JIGSAWS we achieve state-of-the-art performance.

However, closer examination of the held-out predictions on AIxSuture reveals that throughput (suture count) is a particularly strong signal. Although throughput is a valid indicator of surgical performance, its prominence may influence which aspects of skill the model learns to prioritise. If the objective is to develop reliable tools that provide detailed insights into surgical performance, it may be necessary to actively mitigate the influence of throughput and other potentially confounding variables, such as gesture duration. This would help establish whether the model captures broader aspects of technical proficiency beyond speed and efficiency.

The development of robust and generalisable surgical skill assessment models will ultimately require larger datasets, more expert annotations, and longer recordings. The limited availability of expert-level examples is particularly problematic, as the model has fewer opportunities to learn the characteristics of high-level performance. Furthermore, lack of consensus on high-level performance is reflected in the difficulty of separating levels 4 and 5.

Despite these limitations, this work introduces an alternative formulation of surgical skill assessment that explicitly accounts for the heterogeneous distribution of relevant information within a video. By moving beyond uniform clip-level supervision, Surg-Pico provides a foundation for more robust and interpretable approaches to automated surgical assessment.

\section{Methods}\label{sec:methods}
\subsection{Datasets}\label{sec:dataset}

\textbf{JIGSAWS}~\cite{jigsaw} comprises 103 trials of three bench-top
robot-assisted tasks: suturing, needle passing and knot tying. Each video is
rated on six OSATS criteria (respect for tissue, time and motion, overall
performance, suture and needle handling, flow of operation, and quality of the
final product) on a five-point scale. 

\textbf{AIxSuture}~\cite{Hoffmann2024} comprises 314 videos of a suturing task
performed on a bench-top model by medical students, recorded before and after
training. Each video is rated independently by three raters on eight OSATS
criteria (respect for tissue, time and motion, instrument handling, flow of
operation, suture technique, final quality, knowledge of procedure, and overall
performance), plus their sum as a global rating score
(GRS). The OSS challenge defines each OSATS score as the mean of the
three raters. The dataset is substantially imbalanced: the top score accounts
for only 1-9\% of labels across OSATS criteria.

\subsection{Problem formulation}\label{sec:problem}
We represent a video as a bag $\bag = \{\mathbf{x}_1,\dots,\mathbf{x}_M\}$ of fixed-length
clips drawn from it. Each bag carries a single label $\cand^{(e)}_b$ per OSATS criterion $e$. In
PLL, that label is a candidate set $\cand \subseteq \{1,\dots,\nlev\}$ and the
model must recover $y^\star \in \cand$ during training. We formalise the
dataset under the standard PLL assumption as
\begin{equation}
  \mathcal{D} = \bigl\{(\bag_b, \cand_b^{(e)})\bigr\}_{b=1}^{N},
  \qquad e = 1,\dots,\naxes,
  \qquad y_{bm}^{\star(e)} \in \cand_b^{(e)},
  \label{eq:dataset}
\end{equation}
where each clip label \(y_{bm}^{\star(e)}\) needs to be recovered from \(\cand_b^{(e)}\), with $\nlev = 5$ ordinal levels, \(N\) the number of videos, $\naxes = 6$ OSATS criteria on JIGSAWS and
$\naxes = 8$ on AIxSuture.
This follows the homogeneous-bag assumption of MIL~\cite{Carbonneau2018}, where each clip inherits $\cand$,
which increases supervision at no extra annotation cost. Following PiCO+, we relax this assumption by allowing clips for which
\begin{equation}
    y_{bm}^{\star(e)} \notin \cand_b^{(e)},
\end{equation}
so that representations inconsistent with their annotation can be supervised
differently.

\subsection{Architecture}\label{sec:architecture}
A video prediction is built through its global clip centroid (Fig.~\ref{fig:architecture}), where the model aims to project each video clip into a relevant cluster, representative of its underlying skill level.
Clips are independently encoded by a pretrained backbone (Sec.~\ref{sec:model}; Fig.~\ref{fig:architecture}).
The latent representation then branches into two paths: a projection head and
an instance classifier. While the backbone is shared across all OSATS criteria,
we maintain $\naxes$ projection and instance classification heads. We refer to
this model as the query network; unless specified otherwise, the remaining
description applies to a single criterion. A momentum-updated copy of the
network (EMA) is maintained as the key network. A first-in, first-out (FIFO) queue stores each key sample
together with its instance classification prediction, in order to form the
positive sets (see $\mathcal{L}_{\mathrm{cont}}$ in Eq.~\eqref{eq:pico-plus-loss}). Each head maintains its own
candidate sets, prototype bank, pseudo-labels and clean/noisy split; nothing
is shared across OSATS criteria except the representation itself.

\subsection{Objective function}\label{sec:bag-loss}
The full objective is defined as
\begin{equation}
\mathcal{L}
= \frac{1}{N\,\naxes}\sum_{b=1}^{N}\sum_{e=1}^{\naxes}
   \Bigl[\mathcal{L}^{(e)}_{\mathrm{pico+}}(\bag_b) + \mathcal{L}^{(e)}_{\mathrm{bag}}(\bag_b)\Bigr]
 + \lambda_{\mathrm{nce}}\,\frac{1}{N}\sum_{b=1}^{N}\mathcal{L}_{\mathrm{nce}}(\bag_b),
\label{eq:objective}
\end{equation}
where $\mathcal{L}_{\mathrm{pico+}}$ separates a bag into clean ($\mathcal{C}$) and
noisy ($\mathcal{N}$) clips, weighted by $\alpha$ and $\beta$:
\begin{equation}
\mathcal{L}_{\mathrm{pico+}} = \alpha\,\mathcal{L}_{\mathcal{C}} + \beta\,\mathcal{L}_{\mathcal{N}}.
\end{equation}
In our
setting, two raters may assign different performance scores to the same
trial, e.g.\ $\text{Perf} \in \{\text{Medium}, \text{High}\}$. Through a
self-reinforcing loop, PiCO+ allows the model to update the pseudo-label $\pseudo_m$ of
that sample towards what it believes to be the ground truth.
Pseudo-labels are updated during training as
\begin{equation}
    \pseudo_m \leftarrow \phi\, \pseudo_m + (1-\phi)\, \mathbf{z}_m,
    \qquad
    z_{m,c} =
    \begin{cases}
    1 & \text{if } c = \operatorname*{argmax}_{j\in \cand} \hat{\mathbf{q}}_m^\top\proto{j},\\
    0 & \text{otherwise},
    \end{cases}
    \label{eq:pseudo-update}
\end{equation}
where $\hat{\mathbf{q}}_m$ is the clip's normalised projection and $\mathbf{z}_m$ is the one-hot vector of the closest admissible prototype.
As in PiCO+, this update is applied to every clip, clean or noisy, so that a
clip can return to the clean set when the split is recomputed. This is one of the core components of the self-reinforcing loop that drives PiCO+ (Fig.~\ref{fig:architecture}).
\subsubsection{\texorpdfstring{$\mathcal{C}$ and $\mathcal{N}$ terms}{C and N terms}}
Each clip is assigned a prediction $\tilde{y}_m$ by the instance classifier restricted to its candidate set \(Y\). Given the current instance classifier prediction $\tilde{y}_m$, clips are ranked within their video by the cosine similarity between their projection $\hat{\mathbf{q}}_m$ and the corresponding class prototype $\proto{\tilde{y}_m}$.

The hyperparameter $\delta$ defines a threshold $\kappa_\delta$ corresponding to the $(1-\delta)$-quantile of these similarities within the bag, allowing the model to separate the current bag's clips into a clean and a noisy set:
\begin{equation}
    \mathcal{C} = \bigl\{ m \mid \hat{\mathbf{q}}_m^\top\proto{\tilde{y}_m} > \kappa_{\delta}\bigr\},
    \qquad
    \mathcal{N} = \{1,\dots,M\}\setminus\mathcal{C}.
    \label{eq:clean-split}
\end{equation}
The split is recomputed once per epoch. Unlike PiCO+, which ranks samples
across the whole dataset, we take the quantile within each bag, so that every
video contributes a fraction $\delta$ of clean clips.

Prototypes are updated as
\begin{equation}
    \proto{c} \leftarrow \operatorname{Normalise}\bigl(\gamma\,\proto{c} + (1-\gamma)\, \hat{\mathbf{q}}_m\bigr)
    \quad \text{if } c = \tilde{y}_m \text{ and } m \in \mathcal{C},
    \label{eq:proto-update}
\end{equation}
that is, each prototype moves towards the clips the instance classifier
currently assigns to its class, using clean clips only.
PiCO+ then applies different losses to the two sets:
\begin{equation}
\begin{gathered}
    \mathcal{L}_{\mathcal{C}} = \mathcal{L}_{\mathrm{cls}} + \lambda\,\mathcal{L}_{\mathrm{cont}}, \\
    \mathcal{L}_{\mathcal{N}} = \mathcal{L}_{\mathcal{N}\text{-}\mathrm{cont}}
    + \mathcal{L}_{\mathrm{knn}} + \mathcal{L}_{\mathcal{N}\text{-}\mathrm{cls}},
\end{gathered}
\label{eq:pico-plus-loss}
\end{equation}

$\mathcal{L}_{\mathrm{cls}}$ optimises the instance classifier against the
clip's current pseudo-label $\pseudo_m$ through cross-entropy
whereas $\mathcal{L}_{\mathcal{N}\text{-}\mathrm{cls}}$ is
scored against a soft target consisting of the clip's similarity to every
prototype, normalised by a softmax over all $\nlev$ levels. This target is not restricted to the
candidate set, so a noisy clip can be pulled towards any level, including one
outside the raters' ballots; it is recomputed from the current prototypes at
every step and is never written back into $\pseudo_m$. For more details, see Supplementary Note~2.

$\mathcal{L}_{\mathrm{cont}}$ is a standard MoCo/SupCon-style~\cite{He2020moco,Khosla2020supcon} contrastive term
between the current projections $\hat{\mathbf{q}}$ and the projections
$\hat{\mathbf{k}}$ held in a FIFO queue, where positive sets are built from the
candidate-restricted prediction $\tilde{y}_m$ (Fig.~\ref{fig:architecture}).
$\mathcal{L}_{\mathcal{N}\text{-}\mathrm{cont}}$ differs only by relaxing the
prediction to all levels, and $\mathcal{L}_{\mathrm{knn}}$ by using the
$n_{\mathrm{nn}}$ nearest neighbours of each noisy clip as positives. Full
definitions of these terms are given in Supplementary Note~2.

\subsubsection{Bag loss}
We supervise the trial \emph{centroid} in the projection space, where each clip
contributes equally to a global video representation, and decode it
with the prototype bank
$\boldsymbol{\mu} = [\boldsymbol{\mu}_1 \cdots \boldsymbol{\mu}_K] \in \mathbb{R}^{d \times K}$:
\begin{equation}
    \mathbf{c} = \frac{\tfrac{1}{M}\sum_{m=1}^{M} \hat{\mathbf{q}}_m}
                      {\big\lVert \tfrac{1}{M}\sum_{m=1}^{M} \hat{\mathbf{q}}_m \big\rVert_2},
    \qquad
    \tilde{\mathbf{p}} = \operatorname{softmax}\!\left(
        \frac{\mathbf{c}^{\top}\boldsymbol{\mu}}{\tau}\right),
\label{eq:centroid}
\end{equation}
\begin{equation}
    \mathcal{L}_{\mathrm{bag}}
    = -\log \tilde{p}_{y},
    \label{eq:centroid-loss}
\end{equation}
where $y$ is the trial's annotated level: the single annotator's score on
JIGSAWS and the rounded three-rater mean on AIxSuture. The bag term is supervised by a point label rather than
by the candidate set, and is computed over all $M$ clips irrespective of the
clean/noisy partition. Gradients reach $\mathbf{q}$ only: $\boldsymbol{\mu}$ is an EMA
buffer updated without gradients, so bag supervision moves embeddings towards
the correct prototype without corrupting prototype estimation.

\subsubsection{Regularization}
In all experiments, we add an InfoNCE regularisation term~\cite{oord2018cpc,He2020moco}
on the normalised shared trunk latent
$\hat{\mathbf{h}}_m = \mathbf{h}_m/\lVert\mathbf{h}_m\rVert_2$, whose positive is
the key network's latent $\hat{\mathbf{h}}^{k}_m$ for the same clip under the
second augmentation, and whose negatives are the key latents of every clip of
every other trial; clips of the same trial are excluded. Its definition and
weighting are given in Supplementary Note~6.

\subsection{Candidate sets}\label{sec:candidate}
The method above is applied identically to JIGSAWS and AIxSuture; the two differ only in how the candidate set $\cand$ is defined.

\textbf{On JIGSAWS}, we restrict $\cand$ to the video label, which makes
Eq.~\eqref{eq:pseudo-update} inert.
Propagating the trial label to every clip
increases the number of supervised targets ${\approx}18\times$ at no annotation
cost; however, it is detrimental for any clip that does not display the scored behaviour. Our method lets the model decide which clips to trust through the per-epoch selection of Eq.~\eqref{eq:clean-split}. $\delta$ is a hypothesis about how much of a trial carries
scorable signal (Sec.~\ref{sec:model}).

\textbf{On AIxSuture}, the homogeneous-bag assumption is weaker because the videos are
longer: five minutes against ninety seconds for JIGSAWS. The disambiguation
path allows clips to drift away from their instance supervision.
We take the ordinal hull
of the three raters' ballots as the candidate set:
\begin{equation}
  \cand_b
  = \Bigl\{\, i \in \{1,\dots,\nlev\} \;:\;
      \min_{r} y^{(r)}_b \;\le\; i \;\le\; \max_{r} y^{(r)}_b \,\Bigr\},
  \label{eq:cand}
\end{equation}
where $y^{(r)}_b \in \{1,\dots,\nlev\}$ is rater $r$'s score on video $b$ for
the criterion considered. Taking the hull rather than the union of the ballots
avoids asserting that an interior level is excluded: under a union, ballots
$(1,1,3)$ would claim ``1 or 3, but definitely not 2'', for which an ordinal
scale gives no warrant.

The per-clip pseudo-label is initialised from the ballots themselves, at the
measured vote proportions:
\begin{equation}
  s_{m,i}^{(0)} \;=\; \frac{1}{R}\,
    \bigl|\{\, r : y^{(r)}_b = i \,\}\bigr|,
  \qquad m = 1,\dots,M_b,
  \label{eq:sinit}
\end{equation}
where $R$ is the number of raters (on JIGSAWS, $R = 1$, so $\pseudo^{(0)}_m$ is one-hot on $y$). In practice, a 2-1 split starts at
$(0.67, 0.33)$ rather than $(0.5, 0.5)$, and disambiguation must earn any
departure from what the raters said.

Equations~\eqref{eq:cand} and~\eqref{eq:sinit} carry different information,
and the asymmetry is deliberate: $\cand_b$ states what is possible, $\pseudo^{(0)}_m$
what is believed. A level admitted only by the hull is therefore possible but
unsupported by any rater: it starts with zero mass and is reached only if the embedding
supports it.

\subsection{Model and training configuration}\label{sec:model}
A Video Swin-T~\cite{Liu2022videoswin} pretrained on Kinetics-400~\cite{Kay2017kinetics} encodes non-overlapping clips of 10 frames sampled at
2\,fps, at $224\times224$ resolution. It is kept frozen except for LoRA~\cite{Hu2022lora}
adapters of rank 16 with a scaling factor of 32, regularised with a dropout of
0.3. The 768-dimensional clip feature passes through a trunk $h$ (linear layer,
LayerNorm, GELU, linear layer) to 256 dimensions. Two heads read that latent: a
per-criterion instance classifier (linear to 64, LayerNorm, ReLU, dropout 0.3,
linear to $\nlev=5$) and a projector consisting of a shared linear layer with
GELU followed by one linear head per criterion projecting the latent into a
128-dimensional space. The bag centroid is decoded against the prototypes, with
a label smoothing of 0.1 in the loss (Supplementary Note~4). To prevent overfitting, a quarter of each
bag's clips are dropped at random during training.

For the first 10 epochs, the pseudo-labels $\pseudo_m$ are frozen; afterwards they are
updated with a momentum $\phi$ annealed linearly from 0.95 to 0.80 over the
remaining epochs. The FIFO queue holds 8192 keys. The EMA encoder is updated with momentum 0.999.
All temperatures (the contrastive terms, the prototype soft target and the
centroid read-out) are set to $\tau = 0.07$. The contrastive term is weighted
by $\lambda = 0.5$; it is switched on at epoch 1 and ramped to full weight over
10 epochs.

Before epoch 10 no clip is routed to the noisy set ($\mathcal{C}$ holds every
clip and $\mathcal{L}_{\mathcal{N}} = 0$), so the objective is
$\alpha\,\mathcal{L}_{\mathcal{C}}$ computed against the initial pseudo-labels
$\pseudo^{(0)}_m$, and the prototypes are updated with momentum $\gamma = 0.99$ (Eq.~\eqref{eq:proto-update}) from all clips; this is the
warm-up of PiCO+, which runs the supervised contrastive learning on the initial labels before the first
split. The bag loss is enabled at epoch 10 and $\mathcal{L}_{\mathrm{knn}}$ at
epoch 12.

From epoch 10, each bag is partitioned into a clean and a noisy branch,
weighted by $\alpha = 2$ and $\beta = 0.1$, with $n_{\mathrm{nn}} = 32$. $\alpha$, $\beta$, and $\lambda$ were taken from the PiCO+ paper. 
Following PiCO+, which sets $\delta$ according to the expected proportion of
corrupted examples ($\delta = 0.8$ at 10\% noise, down to $0.5$ at 40\%), we
fixed $\delta$ a priori rather than tuning it. AIxSuture trials are a
continuous suturing exercise, so we used $\delta = 0.8$. On JIGSAWS, the core
gestures (G2-G4 for suturing and needle passing; G13-G15 for knot tying)
account for 63\% of each trial on average in the gesture annotations, leaving
about 37\% of each trial outside them; we therefore used $\delta = 0.5$, the
PiCO+ setting for comparable noise.

Optimisation uses SGD with
momentum 0.9, a learning rate of $5\times10^{-4}$ and a weight decay of
$10^{-3}$ (excluding norm and bias parameters), under a cosine schedule with
five warm-up epochs. Training runs for a fixed 100
epochs at a batch size of 2. Augmentations are applied consistently across the frames of each clip, with fewer operations for the key view than for the query view. Details can be found in Supplementary Note~5.
Held-out metrics are the mean over the final 5 epochs, without
checkpoint selection. Every experiment was run with three different seeds. We report their mean $\pm$ SD.

\subsection{Cross-validation}\label{sec:protocol-cv}
On JIGSAWS, we use leave-one-supertrial-out (LOSO) cross-validation for
comparability with prior work on this benchmark, with leave-one-user-out
(LOUO) as a control. On AIxSuture, we adopt two evaluation protocols. The first
is the official 70/15/15 training, validation and test split of the dataset
paper~\cite{Hoffmann2024} where the checkpoint with the lowest validation EC is evaluated on the test set. To our knowledge, no previous work reports per-criterion OSATS results on the open-source data alone. We include results from the recent OSS
challenge; however, the challenge used additional private data and different
folds, which makes direct comparison difficult. We therefore also report a
4-fold leave-multiple-users-out (LMUO) cross-validation, where the training and test folds do not contain overlapping participant IDs, matching the LOUO protocol on JIGSAWS, to
establish a baseline on this dataset. Cross-validation is also the only
protocol that allows the model to be compared against the label-reliability
ceiling across the whole corpus.

\subsection{Metrics}\label{sec:protocol-metrics}
We report Spearman's $\rho$ per criterion and averaged over criteria
($\rho_{\mathrm{OSATS}}$) and $\rho$ on the global rating score. For
comparability with the OSS challenge, we additionally report its two scoring
metrics: macro-F1, computed per OSATS criterion over all available $\nlev$
classes and averaged across criteria, and the expected cost
$\mathrm{EC} = \frac{1}{N}\sum_{i}\sum_{j} w_{ij} n_{ij}$, where $n_{ij}$ is the
number of videos of level $i$ predicted as level $j$, $N$ the number of videos
and $w_{ij} = |i-j|/(\nlev-1)$.  At test time, the score of a video on criterion $e$
is the expected level under the centroid softmax of Eq.~\eqref{eq:centroid},
$\hat{y}^{(e)} = \sum_{k=1}^{\nlev} k\,\tilde{p}^{(e)}_k$. Spearman's $\rho$
is computed on $\hat{y}^{(e)}$, and macro-F1 and EC on the argmax of
$\tilde{\mathbf{p}}^{(e)}$. GRS is reported on AIxSuture and is the sum of the criterion
scores, $\sum_{e=1}^{\naxes}\hat{y}^{(e)}$; GRS is never regressed directly. Performance metrics are averaged across folds,
as in ReCAP~\cite{recap}.
On the official split, level 5 appears only in the test or validation split for
some criteria. We therefore cap the labels at 4 for this experiment, mapping every level-5 video to
level 4. As reducing the number of classes inflates both F1 and EC, these two
metrics are not comparable with previous work; we report them so
that future work can compare against the official split
(Table~\ref{tab:aix-main}).

\subsection{Label reliability on AIxSuture}\label{sec:protocol-ceiling}
Each AIxSuture video is scored by three raters, and models are evaluated
against the mean of their scores. Because the raters disagree, this mean is
itself a noisy estimate of the true skill level, which limits how well any
model can correlate with it. Classical test theory makes this precise: if the
evaluation target $Y_{\mathrm{obs}}$ measures the true level $Y$ with
reliability $\rho_{YY}$, then for any prediction $\hat{Y}$~\cite{Spearman1904-hx}
\begin{equation}\label{eq:attenuation}
  \operatorname{corr}(\hat{Y}, Y_{\mathrm{obs}})
    = \operatorname{corr}(\hat{Y}, Y)\,\sqrt{\rho_{YY}}
    \;\le\; \sqrt{\rho_{YY}}.
\end{equation}
Even a model that recovers $Y$ exactly therefore scores at most
$\sqrt{\rho_{YY}}$, which we refer to as the ceiling. We take $\rho_{YY}$ to be
the reliability of the three-rater mean, the consistency, average-measures
intraclass correlation $\mathrm{ICC}(C,3)$~\cite{McGrawWong1996,ShroutFleiss1979}.
The ceiling differs by criterion, so every correlation reported in
Sec.~\ref{sec:results} is accompanied by its own criterion's bound; the full
reliability analysis is given in Supplementary Note~1.

\backmatter

\bmhead{Data availability}
This work uses two public datasets and generates no new data.
JIGSAWS~\cite{jigsaw} and AIxSuture~\cite{Hoffmann2024} are available from their
respective custodians under their own data-use agreements.

\bmhead{Code availability}
The code for this project is accessible on GitHub
(\url{https://github.com/JK-rez/Surg-Pico}).

\bmhead{Acknowledgements}
This work was supported by core funding from the Wellcome/EPSRC Centre for
Medical Engineering [WT203148/Z/16/Z]. For the purpose of Open Access, the
authors have applied a CC BY public copyright licence to any Author Accepted
Manuscript version arising from this submission.

\bmhead{Author contributions}
J.Q. implemented and ran all experiments. J.Q. and A.G. contributed to data interpretation, method design, critical
review, writing and revision of the manuscript. M.M. contributed to the method design. S.O., J.S. and A.G. supervised the project and acquired the
funding. All authors have read and approved the final version of the manuscript.

\bmhead{Competing interests}
The authors declare no competing interests.

\bmhead{Use of large language models}
A large language model (Claude, Anthropic) was used to edit and format the
manuscript, to draft text that the authors then revised, to code the supporting analyses of rater agreement, and to draft figure layouts that the authors then edited. The authors
reviewed and verified all AI-assisted output and take full responsibility for
the scientific content, analyses and conclusions of this work.

\bibliography{references}
\end{document}


\title{Supplementary Information for ``Surg-Pico: Disentangling Granular Surgical Skills from Video-Level Labels''}

\author*[1]{\fnm{Julien} \sur{Quarez}}\email{julien.quarez@kcl.ac.uk}
\author[1]{\fnm{Marc} \sur{Modat}}
\author[1]{\fnm{Sebastien} \sur{Ourselin}}
\author[1,2]{\fnm{Jonathan} \sur{Shapey}}
\author[1]{\fnm{Alejandro} \sur{Granados}}

\affil[1]{\orgdiv{School of Biomedical Engineering and Imaging Sciences},
\orgname{King's College London}, \orgaddress{\city{London}, \country{UK}}}
\affil[2]{\orgdiv{Neurosurgery}, \orgname{King's College Hospital},
\orgaddress{\city{London}, \country{UK}}}

\maketitle

\section*{Supplementary Note 1: Label reliability on AIxSuture}
The main text bounds the correlation any model can reach against the
three-rater mean by $\sqrt{\rho_{YY}}$, where $\rho_{YY}$ is the reliability of
that mean. We take $\rho_{YY}$ to be the consistency, average-measures intraclass
correlation $\mathrm{ICC}(C,3)$~\cite{McGrawWong1996} (ICC(3,$k$)
in~\cite{ShroutFleiss1979}). We use the consistency form because Spearman's
$\rho$ is unaffected by a rater who scores everyone slightly higher or lower;
the absolute-agreement form would count such offsets as error and understate
the ceiling. The attenuation bound is exact for Pearson's correlation, and we
use it as an approximate bound for Spearman's $\rho$.

For the global rating score, $\mathrm{ICC}(C,3) = 0.928$ ($95\%$ CI
$[0.914, 0.941]$, $n = 314$), giving a ceiling of $0.963$. As a check, the
mean pairwise inter-rater correlation ($r = 0.816$), extended to three raters
with the Spearman--Brown formula $3r/(1+2r)$, gives $0.930$; the pairwise
correlation is in line with the average of about $0.8$ reported for this
corpus~\cite{Hoffmann2024}.
Supplementary Table~\ref{tab:ceiling} reports the same quantities for each
OSATS criterion.

\begin{table}[h]
\centering
\caption{Label reliability and the resulting ceiling on AIxSuture, over all
$n = 314$ videos rated by three raters. $\mathrm{ICC}(C,3)$ is the reliability
of the three-rater mean, the target the model is scored against; the ceiling
is $\sqrt{\mathrm{ICC}(C,3)}$ (attenuation bound, main text).
$\mathrm{ICC}(C,1)$ is the corresponding single-rater reliability, which
ranges from $0.54$ to $0.81$ across criteria.
The ceiling spans eight points across criteria, so a correlation is only
interpretable beside its own bound.}
\label{tab:ceiling}
\begin{tabular}{lcccc}
\toprule
OSATS criterion & $\mathrm{ICC}(C,3)$ & $95\%$ CI & $\mathrm{ICC}(C,1)$ & Ceiling \\
\midrule
Respect for Tissue      & 0.778 & [0.731, 0.817] & 0.538 & 0.882 \\
Time and Motion         & 0.852 & [0.821, 0.878] & 0.657 & 0.923 \\
Instrument Handling     & 0.855 & [0.825, 0.881] & 0.664 & 0.925 \\
Suture Technique        & 0.873 & [0.847, 0.895] & 0.696 & 0.934 \\
Flow of Operation       & 0.849 & [0.818, 0.876] & 0.653 & 0.921 \\
Knowledge of Procedure  & 0.926 & [0.910, 0.939] & 0.806 & 0.962 \\
Overall Performance     & 0.911 & [0.892, 0.926] & 0.772 & 0.954 \\
Final Quality           & 0.906 & [0.887, 0.923] & 0.764 & 0.952 \\
\midrule
Global Rating Score     & 0.928 & [0.914, 0.941] & 0.812 & 0.963 \\
\bottomrule
\end{tabular}
\end{table}

Because the ceiling is a property of the target rather than of the model, it
differs by criterion (Supplementary Table~\ref{tab:ceiling}). Respect for
tissue is the least reliably rated criterion ($\mathrm{ICC}(C,3) = 0.778$,
ceiling $0.882$) and knowledge of procedure the most reliably rated ($0.926$,
ceiling $0.962$). A model scoring $\rho = 0.80$ therefore reaches $91\%$ of the
attainable correlation on the former and $83\%$ on the latter, and an
aggregate figure quoted across criteria conceals that spread.

The consistency and absolute-agreement forms differ by less than $0.03$ on
every criterion except instrument handling, where $\mathrm{ICC}(A,3) = 0.779$
against $\mathrm{ICC}(C,3) = 0.855$. The raters therefore rank trainees
similarly on instrument handling but differ systematically in where they place
them on the scale; it is the one criterion on which an absolute score is less
transferable between raters than a relative one.

\section*{Supplementary Note 2: PiCO+ components}
We give the full definitions of the PiCO+~\cite{Wang2022PiCOPlus} components
that Surg-Pico builds on, in the notation of the main text (Methods, Objective
function). For one OSATS criterion, each clip $\mathbf{x}_m$ of a bag has a
candidate set $\cand$, an instance classifier output $f(\mathbf{x}_m)$, a
projection $\mathbf{q}_m$ and a maintained pseudo-label $\pseudo_m$; one
unit-norm prototype $\proto{c}$ is kept per level.

\paragraph{Normalised projections.}
Projections are $\ell_2$-normalised,
\begin{equation}
    \hat{\mathbf{q}}_m = \frac{\mathbf{q}_m}{\lVert \mathbf{q}_m \rVert_2},
    \label{eq:s-norm}
\end{equation}
so that inner products with the prototypes are cosine similarities. The same
normalisation is applied to every clip before the bag centroid is formed.

\paragraph{Classification loss and the disambiguation loop.}
The instance classifier is trained against the pseudo-label by soft
cross-entropy,
\begin{equation}
    \mathcal{L}_{\mathrm{cls}}(\mathbf{x}_m) = -\sum_{j=1}^{\nlev} s_{m,j} \log f_j(\mathbf{x}_m).
    \label{eq:s-cls}
\end{equation}
The pseudo-label is moved towards the closest admissible prototype, and each
prototype towards the clean clips the classifier assigns to its level (main
text, pseudo-label and prototype updates). This is a self-reinforcing loop:
prototypes are built from the classifier's assignments, and the classifier's
target is set by the prototypes. Neither is optimised against a fixed target,
so the loop is bounded only by the candidate set, which makes the choice of
candidate set the central design decision. The pseudo-label update is applied
to every clip, clean or noisy, because the clean/noisy split is recomputed
every epoch and a clip must be able to return to the clean set.

\paragraph{Contrastive term.}
$\mathcal{L}_{\mathrm{cont}}$ is a MoCo/SupCon-style contrastive term between
query projections $\hat{\mathbf{q}}$ and momentum-encoder key projections
$\hat{\mathbf{k}}$. Writing $\mathcal{V}_q$ and $\mathcal{V}_k$ for the query
and key views of the current batch and $\mathcal{Q}$ for the FIFO queue of keys
from previous batches, the contrastive pool of clip $m$ excludes the anchor
itself,
\begin{equation}
    \mathcal{A}_m = \bigl(\mathcal{V}_q \cup \mathcal{V}_k \cup \mathcal{Q}\bigr) \setminus \{\hat{\mathbf{q}}_m\},
    \label{eq:s-pool}
\end{equation}
and its positives are the pool elements given the same candidate-restricted
prediction $\tilde{y}_m = \operatorname*{argmax}_{j\in \cand} f_j(\mathbf{x}_m)$:
\begin{equation}
    P_m = \bigl\{\, \mathbf{v} \in \mathcal{A}_m \;:\; \tilde{y}(\mathbf{v}) = \tilde{y}_m \,\bigr\},
    \label{eq:s-pos}
\end{equation}
\begin{equation}
    \mathcal{L}_{\mathrm{cont}}(\mathbf{x}_m) = -\frac{1}{|P_m|}\sum_{\mathbf{v}^{+}\in P_m}
    \log \frac{\exp(\hat{\mathbf{q}}_m^\top \mathbf{v}^{+}/\tau)}
              {\sum_{\mathbf{v}\in \mathcal{A}_m}\exp(\hat{\mathbf{q}}_m^\top \mathbf{v}/\tau)}.
    \label{eq:s-cont}
\end{equation}
Only clean clips update the
prototypes. PiCO+ selects the clean set by ranking samples across the whole
training set; Surg-Pico ranks the clips within each video instead (main text,
Methods), so that every video contributes the same fraction of clean clips.

\paragraph{Noisy-set terms.}
Noisy clips are trained with three terms.
$\mathcal{L}_{\mathcal{N}\text{-}\mathrm{cont}}$ takes the form of
Eq.~\eqref{eq:s-cont}, but its positive set uses the unrestricted prediction
for noisy clips,
\begin{equation}
    \tilde{y}_m =
    \begin{cases}
        \operatorname*{argmax}_{1 \leq j \leq \nlev} f_j(\mathbf{x}_m) & \text{if } m \in \mathcal{N}, \\
        \operatorname*{argmax}_{j \in \cand} f_j(\mathbf{x}_m) & \text{otherwise.}
    \end{cases}
    \label{eq:s-noisy-pred}
\end{equation}
$\mathcal{L}_{\mathrm{knn}}$ also takes the form of Eq.~\eqref{eq:s-cont}, with
the $n_{\mathrm{nn}}$ nearest neighbours of the noisy clip in $\mathcal{A}_m$ as
positives. $\mathcal{L}_{\mathcal{N}\text{-}\mathrm{cls}}$ is a soft
cross-entropy against a target computed from the prototypes over all levels,
\begin{equation}
    s'_{m,j} = \frac{\exp(\hat{\mathbf{q}}_m^\top\proto{j}/\tau)}{\sum_{t=1}^{\nlev}\exp(\hat{\mathbf{q}}_m^\top\proto{t}/\tau)},
    \qquad 1\leq j\leq \nlev.
    \label{eq:s-soft-target}
\end{equation}
Unlike $\pseudo_m$, $\mathbf{s}'_m$ is computed afresh at each step, is not
restricted to the candidate set, and is never written back into $\pseudo_m$.

\section*{Supplementary Note 3: ReCAP formulation}
ReCAP~\cite{recap} is the direct ancestor of Surg-Pico. It was introduced on the
kinematic signals of JIGSAWS: a recurrent cross-attention network predicts
OSATS scores for consecutive segments of a trial, the trial prediction is the
average of these segment predictions, and only that average is supervised by
the trial label. The segment predictions can then be read back as pseudo-labels
at no extra annotation cost.

The ReCAP read-out of the ablation in the main text keeps this aggregation and
applies it to the video clips of a bag. For OSATS criterion $e$, the latent
$\mathbf{h}_m$ of clip $\mathbf{x}_m$ (the trunk output, main text) is mapped
to a distribution over the $\nlev = 5$ levels by the prediction head
\begin{equation}
    f^{(e)}(\mathbf{x}_m) = \operatorname{softmax}\Bigl(\mathbf{W}^{(e)}_2\,
      \operatorname{Dropout}\bigl(\operatorname{ReLU}\bigl(\operatorname{LN}
      (\mathbf{W}^{(e)}_1 \mathbf{h}_m + \mathbf{b}^{(e)}_1)\bigr)\bigr)
      + \mathbf{b}^{(e)}_2\Bigr),
    \label{eq:s-head}
\end{equation}
with a 64-unit hidden layer, layer normalisation (LN) and a dropout of 0.3.
This is the same instance classifier $f$ that Surg-Pico trains with
$\mathcal{L}_{\mathrm{cls}}$ (Supplementary Note~2). The bag prediction is the
uniform mixture of the per-clip distributions,
\begin{equation}
    \bar{\mathbf{p}}^{(e)} = \frac{1}{M}\sum_{m=1}^{M} f^{(e)}(\mathbf{x}_m),
    \label{eq:s-recap-bag}
\end{equation}
trained by cross-entropy against the trial level $y^{(e)}$ with label
smoothing $\varepsilon = 0.1$,
\begin{equation}
    \mathcal{L}_{\mathrm{ReCAP}}
    = -(1-\varepsilon)\log \bar{p}^{(e)}_{y^{(e)}}
      - \frac{\varepsilon}{\nlev}\sum_{k=1}^{\nlev}\log \bar{p}^{(e)}_{k}.
    \label{eq:s-recap-loss}
\end{equation}
The per-clip expected level $\sum_{k} k\, f^{(e)}_k(\mathbf{x}_m)$ provides the
interpretability trace. Compared with the original head, layer normalisation
replaces batch normalisation, because a batch holds the clips of only two
trials and batch statistics would leak trial context into the per-clip
predictions, and the label smoothing is lowered from 0.3 to 0.1.

Because only the mixture $\bar{\mathbf{p}}^{(e)}$ is supervised, every clip is
treated as an equally valid observation of the trial's skill, which is the
assumption the clean/noisy selection of Surg-Pico relaxes.

\section*{Supplementary Note 4: Bag loss with label smoothing}
The bag loss of the main text is trained with label smoothing
$\varepsilon = 0.1$:
\begin{equation}
    \mathcal{L}_{\mathrm{bag}}
    = -(1-\varepsilon)\log \tilde{p}_{y}
      - \frac{\varepsilon}{\nlev}\sum_{i=1}^{\nlev}\log \tilde{p}_{i},
    \label{eq:s-bag}
\end{equation}
where $\tilde{\mathbf{p}}$ is the softmax read-out of the bag centroid against
the prototypes and $y$ is the trial's level.

\section*{Supplementary Note 5: Data augmentation}
Each clip is augmented twice, once for the query network and once for the key
(momentum) network. Augmentations act on pixel values before normalisation,
and their random parameters are drawn once per clip and shared by all of its
frames, so that motion is preserved.
\begin{itemize}
\item \textbf{Query view:} a horizontal flip, a mild random resized crop that
  keeps most of the field of view (85--100\% of the frame), colour jitter
  (brightness, contrast, saturation and hue), a per-channel white-balance gain,
  a gamma correction and low-amplitude sensor noise.
\item \textbf{Key view:} a horizontal flip followed by two photometric
  operations drawn at random from brightness, contrast, saturation, hue, white
  balance, gamma and blur, and an occasional conversion to grayscale. It applies
  fewer operations than the query view, so that the keys stored in the FIFO
  queue stay close to the clip as recorded.
\end{itemize}
Vertical flips, temporal reversal and random erasing are deliberately excluded:
they would alter or hide the instruments, the needle or the motion on which the
OSATS criteria are scored. At the bag level, a quarter of each bag's clips are
dropped at random at every training step (main text).

\section*{Supplementary Note 6: Regularisation Term}
The PiCO+ and bag terms act on the projection space of each OSATS criterion;
none acts directly on the trunk latent $\mathbf{h}_m$ that all criteria share.
The regularisation term of the main text (Methods, Objective function) is an
InfoNCE loss~\cite{oord2018cpc,He2020moco} on that latent, used in all
experiments, including every configuration of the ablation.

\paragraph{Definition.}
Let $\hat{\mathbf{h}}_m = \mathbf{h}_m/\lVert\mathbf{h}_m\rVert_2$ be the
normalised trunk latent of clip $\mathbf{x}_m$ of bag $\bag_b$ from the query
network, and $\hat{\mathbf{h}}^{k}_m$ the normalised latent of the same clip
under the second augmentation from the key (momentum) network. Key latents are
stored in a memory bank $\mathcal{M}$ with one entry per training clip. The
term is
\begin{equation}
  \mathcal{L}_{\mathrm{nce}} = -\log
  \frac{\exp(\hat{\mathbf{h}}_m^\top\hat{\mathbf{h}}^{k}_m/\tau_{\mathrm{nce}})}
       {\exp(\hat{\mathbf{h}}_m^\top\hat{\mathbf{h}}^{k}_m/\tau_{\mathrm{nce}})
        + \sum_{\mathbf{v}\in\mathcal{M}\setminus\bag_b}\exp(\hat{\mathbf{h}}_m^\top\mathbf{v}/\tau_{\mathrm{nce}})},
  \label{eq:s-nce}
\end{equation}
averaged over the clips of the bag. Gradients reach $\mathbf{h}_m$ only: the
key latents and the bank are computed without gradients. The term acts on the
normalised latent, so it constrains the direction of $\mathbf{h}_m$ and leaves
its norm free.

\paragraph{Same-trial exclusion.}
All clips of the anchor's own trial, $\bag_b$, are removed from the pool and are
neither positives nor negatives. Clips of one trial share the surgeon, the
session, the task and the label; using them as negatives would push apart the
trial-level signal that the bag read-out is built on, and using them as
positives would duplicate the bag supervision. With the trial excluded, the
task the term sets is which trial a clip comes from, a signal the skill labels
do not contain.
\paragraph{Memory bank.}
The bank holds one entry per training clip rather than a FIFO queue, so that
the negatives cover every other trial of the training fold. Entries are
written from the key network immediately after its forward pass and are
normalised when read; an entry is therefore at most one epoch old, measured
against a key network that moves with momentum $0.999$. Entries not yet
written are masked out, and a video is skipped while the bank holds fewer than
64 valid negatives, which happens only during the first epoch.

\paragraph{Weight.}
$\lambda_{\mathrm{nce}}$ is not set by hand, because the trunk latent is
unnormalised and a fixed weight would leave the term either negligible or
dominant. During its first active epoch, the term is computed but not added to
the loss, and $\lambda_{\mathrm{nce}}$ is set so that its gradient on the
latent matches that of all other terms combined, averaged over that epoch:
\begin{equation}
  \lambda_{\mathrm{nce}}\,
  \Bigl\lVert\frac{\partial\mathcal{L}_{\mathrm{nce}}}{\partial\mathbf{h}}\Bigr\rVert
  = r\,\Bigl\lVert\frac{\partial\mathcal{L}_{\mathrm{rest}}}{\partial\mathbf{h}}\Bigr\rVert,
  \qquad r = 1.
  \label{eq:s-nce-parity}
\end{equation}
The weight is then frozen and the term is applied from the following epoch to
the end of training.

\paragraph{Temperature.}
We use $\tau_{\mathrm{nce}} = 0.2$, above the $0.07$ of the supervised terms.
Those terms have many positives per anchor; Eq.~\eqref{eq:s-nce} has one
positive against every clip of every other trial in the fold, where a
temperature of $0.07$ would make the softmax close to a hard maximum over the
bank's slightly stale entries.

\bibliography{references}